%% file: arxiv.tex
\documentclass[10pt,twocolumn,letterpaper]{article}

\usepackage{iccv}
\usepackage{times}
\usepackage{epsfig}
\usepackage{graphicx}
\usepackage{amsmath}
\usepackage{amssymb}

\usepackage{enumitem}
\usepackage{subcaption}
\usepackage{adjustbox}
\usepackage{multirow}
\usepackage{bbding}

\usepackage[pagebackref=true,breaklinks=true,letterpaper=true,colorlinks,bookmarks=false]{hyperref}
\usepackage{xcolor}

\iccvfinalcopy 

\def\iccvPaperID{****} 

\ificcvfinal\fi

\title{PointPatchMix: Point Cloud Mixing with Patch Scoring}

\author{Yi Wang$^{1*}$, Jiaze Wang$^{2*}$, Jinpeng Li$^{2}$, Zixu Zhao$^{2}$, \\ Guangyong Chen$^{3}$, Anfeng Liu$^{1\dag}$ and Pheng-Ann Heng$^{2}$
}

\begin{document}


\maketitle
\def\thefootnote{*}\footnotetext{These authors contributed equally to this work}\def\thefootnote{\arabic{footnote}}
\def\thefootnote{$\dag$}\footnotetext{Corresponding Author: afengliu@mail.csu.edu.cn}\def\thefootnote{\arabic{footnote}}
\def\thefootnote{1}\footnotetext{Computer Science and Engineering, Central South University}\def\thefootnote{\arabic{footnote}}
\def\thefootnote{2}\footnotetext{Computer Science and Engineering, The Chinese University of Hong Kong}\def\thefootnote{\arabic{footnote}}
\def\thefootnote{3}\footnotetext{Zhejiang Lab, Hangzhou, China}\def\thefootnote{\arabic{footnote}}

\ificcvfinal\thispagestyle{empty}\fi

\input{abstract}
\input{introduction}
\input{relatedworks}

\input{method}
\input{experiments}
\input{conclusion}

{\small
\bibliographystyle{ieee_fullname}
\bibliography{egbib}
}

\end{document}


\title{PointPatchMix: Point Cloud Mixing with Patch Scoring}

\author{First Author\\
Institution1\\
Institution1 address\\
{\tt\small firstauthor@i1.org}
\and
Second Author\\
Institution2\\
First line of institution2 address\\
{\tt\small secondauthor@i2.org}
}

\maketitle
\ificcvfinal\thispagestyle{empty}\fi

\section{}

\section{Visuations}

{\small
\bibliographystyle{ieee_fullname}
\bibliography{egbib}
}

%% file: abstract.tex
\begin{abstract}

Data augmentation is an effective regularization strategy for mitigating overfitting in deep neural networks, and it plays a crucial role in 3D vision tasks, where the point cloud data is relatively limited. While mixing-based augmentation has shown promise for point clouds, previous methods mix point clouds either on block level or point level, which has constrained their ability to strike a balance between generating diverse training samples and preserving the local characteristics of point clouds. Additionally, the varying importance of each part of the point clouds has not been fully considered, cause not all parts contribute equally to the classification task, and some parts may contain unimportant or redundant information. To overcome these challenges, we propose PointPatchMix, a novel approach that mixes point clouds at the patch level and integrates a patch scoring module to generate content-based targets for mixed point clouds. Our approach preserves local features at the patch level, while the patch scoring module assigns targets based on the content-based significance score from a pre-trained teacher model. We evaluate PointPatchMix on two benchmark datasets, ModelNet40 and ScanObjectNN, and demonstrate significant improvements over various baselines in both synthetic and real-world datasets, as well as few-shot settings. With Point-MAE as our baseline, our model surpasses previous methods by a significant margin, achieving 86.3\% accuracy on ScanObjectNN and 94.1\% accuracy on ModelNet40. Furthermore, our approach shows strong generalization across multiple architectures and enhances the robustness of the baseline model.

\end{abstract}

%% file: introduction.tex
\section{Introduction}
\begin{figure}[t]
    \centering
        \includegraphics[width=\linewidth]{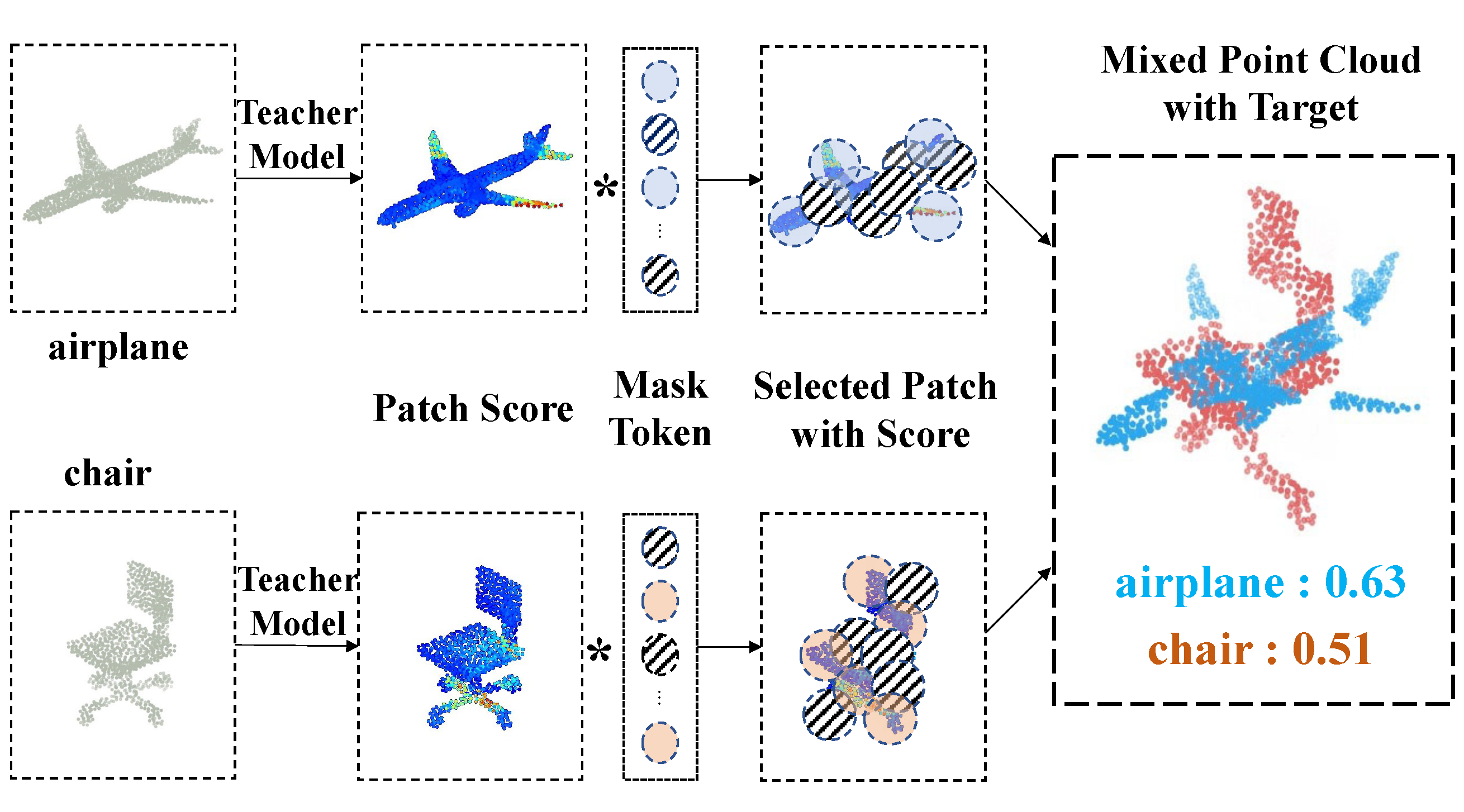}
    \caption{\textbf{Illustration of generating mixed data using PointPatchMix.} Given two point clouds, PointPatchMix processes them at the patch level, with each patch comprising 32 points. A pre-trained teacher model scores each patch based on self-attention mechanism. Then, the mixed point cloud consists of patches selected by mask tokens and the new ground truth is generated.}
    \label{fig:teaser}
    \vspace{-10pt}
\end{figure}

Accurately classifying point clouds has become an increasingly crucial and challenging research topic in the field of computer vision, with extensive applications in real-world scenarios such as virtual reality, augmented reality, and autonomous navigation. Despite the limited availability of point cloud data, deep neural networks (DNNs) have made remarkable progress in this area in recent years ~\cite{qi2017pointnet,qi2017pointnet++,zhao2021point,pang2022masked,yu2022point}. However, DNNs are prone to overfitting because they tend to approximate the model from the given data distribution, particularly when the point cloud training data is scarce. To overcome this challenge, many researchers~\cite{li2020pointaugment,zhang2022pointcutmix,kim2021point} propose the use of point cloud data augmentation techniques to enhance the diversity of the training data and improve the generalizability of the model. 

Many methods have been proposed to augment point cloud data, with both manually designed~\cite{ren2022benchmarking,qi2017pointnet,qi2017pointnet++,goyal2021revisiting} and automatically generated methods~\cite{li2020pointaugment,zhang2022pointcutmix,chen2020pointmixup,lee2021regularization}. Among them, mixing-based augmentation~\cite{zhang2022pointcutmix,chen2020pointmixup,lee2021regularization} has shown great potential for point clouds. Mixing-based augmentation generates labels for mixed point clouds based on their original labels. For instance, PointMixup~\cite{chen2020pointmixup} defines data augmentation between point clouds as a shortest path linear interpolation and generates new samples by optimally assigning a path function. Similarly, PointCutMix \cite{zhang2022pointcutmix} finds the optimal assignment between two point clouds and generates new training samples by replacing points in one sample with their corresponding pairs in the other sample. Moreover, RSMix~\cite{lee2021regularization} generates a virtual mixed sample by replacing part of the sample with shape-preserved subsets from another sample.


Despite the significant advancements in mixing-based techniques, there still remain some limitations that require attention. Firstly, previous methods such as PointMixup~\cite{chen2020pointmixup} and PointCutMix~\cite{zhang2022pointcutmix} generate mixed point clouds at either block or point levels. While point-level mixing can increase the diversity of training samples, it may not adequately preserve the local characteristics of the point cloud, whereas block-level mixing may limit the diversity of the generated point clouds while retaining local features. Striking a balance between diversity and accuracy can therefore be challenging. Moreover, existing methods, such as PointCutMix, generate targets for mixed point clouds through a simple linear combination based on the sample ratio of each point cloud. However, this approach may not be optimal for many scenarios where the importance of each patch in the point cloud may significantly differ. Therefore, it is imperative to develop an augmentation method at the patch level that generates content-based targets.

To overcome these challenges, we propose PointPatchMix, a novel point cloud mixing method that operates at the patch level and integrates a patch scoring module. 
As shown in Figure \ref{fig:teaser}, our patch-level mixing approach generates diverse training samples while preserving the local features of the point clouds, making it an ideal choice for the Transformer architecture, which employs patch embeddings as the basic input units. Additionally, we observe that not all point patches contribute equally to the final classification scores, and some may contain unimportant or redundant information. To address this, we generate significance scores for each patch using a pre-trained teacher network, which considers the value of the information contained in each patch.
We conduct extensive experiments on ModelNet40 and ScanObjectNN, and evaluate PointPatchMix with various baselines. 
With Point-MAE~\cite{pang2022masked} as the baseline, our model achieves 86.3\% accuracy on ScanObjectNN and 94.1\% accuracy on ModelNet40, which outperforms previous methods by a notable margin. PointPatchMix also demonstrates strong generalization across multiple architectures and can well enhance the baseline model's robustness.

Our main contributions can be summarized as follows:
\begin{itemize}[noitemsep,topsep=0pt]
\setlength{\itemsep}{0pt}
\item We introduce PointPatchMix, an innovative patch-level augmentation technique that produces diverse and realistic augmented training samples.
\item A patch scoring module is proposed to generate content-based targets for the mixed point clouds, which allows for more accurate ground truths for classification.
\item Our results demonstrate that PointPatchMix achieves significant improvements in both synthetic datasets, real-world datasets and few-shot settings. 
\end{itemize}

%% file: relatedworks.tex
\section{Related Works}
\noindent\textbf{Transformer with self-attention on point cloud.} 
The transformer model~\cite{vaswani2017attention}, which utilizes the self-attention mechanism, has proven to be highly successful in the field of NLP~\cite{brown2020language,lee2018pre,radford2019language,raffel2020exploring}, outperforming Convolutional Neural Networks (CNNs) and Recurrent Neural Networks (RNNs) and becoming the primary feature processor in the field.
 This achievement has been extended to the two-dimensional image domain with ViT~\cite{dosovitskiy2020image,yuan2021tokens,yuan2022volo,wang2021pyramid,wang2021crossformer,liu2021swin,zhao2021point,fan2021review,fan2022generalized,fan2023learnable}.
 In the point cloud, the transformer architecture is also constantly evolving and improving~\cite{liu2021survey}. 
 The Point transformer~\cite{zhao2021point} uses the self-attention mechanism locally to better apply the transformer architecture to large and complex 3D scenarios. 
 Recently, Point-MAE~\cite{pang2022masked} demonstrates the significant potential of standard transformer in processing point cloud data for self-supervised learning by using masked self-coding scheme and displaying high generalization ability in downstream tasks.

\noindent\textbf{Data augmentation on images.} 
Data augmentation is an effective method for increasing the complexity of a dataset at minimal cost by transforming existing data to help the model learn more knowledge and improve accuracy.
For images, except for traditional methods (i.e. random flipping, cropping, rotation), some works~\cite{zhang2020does,zhang2017mixup,verma2019manifold,walawalkar2020attentive,yun2019cutmix,liu2022tokenmix}
combine two samples to generate a mixed sample with dual labels, significantly improving the difficulty of the sample and further optimizing the model.  
Typical representatives in these works are Mixup~\cite{zhang2017mixup} and CutMix~\cite{yun2019cutmix}. 
MixUp interpolates the pixels of the two images in a certain ratio to create an overlap of the two.
CutMix cuts two images regionally and splices the cut regions to create the mixed sample.

\begin{figure*}[ht]
    \centering
        \includegraphics[width=\linewidth]{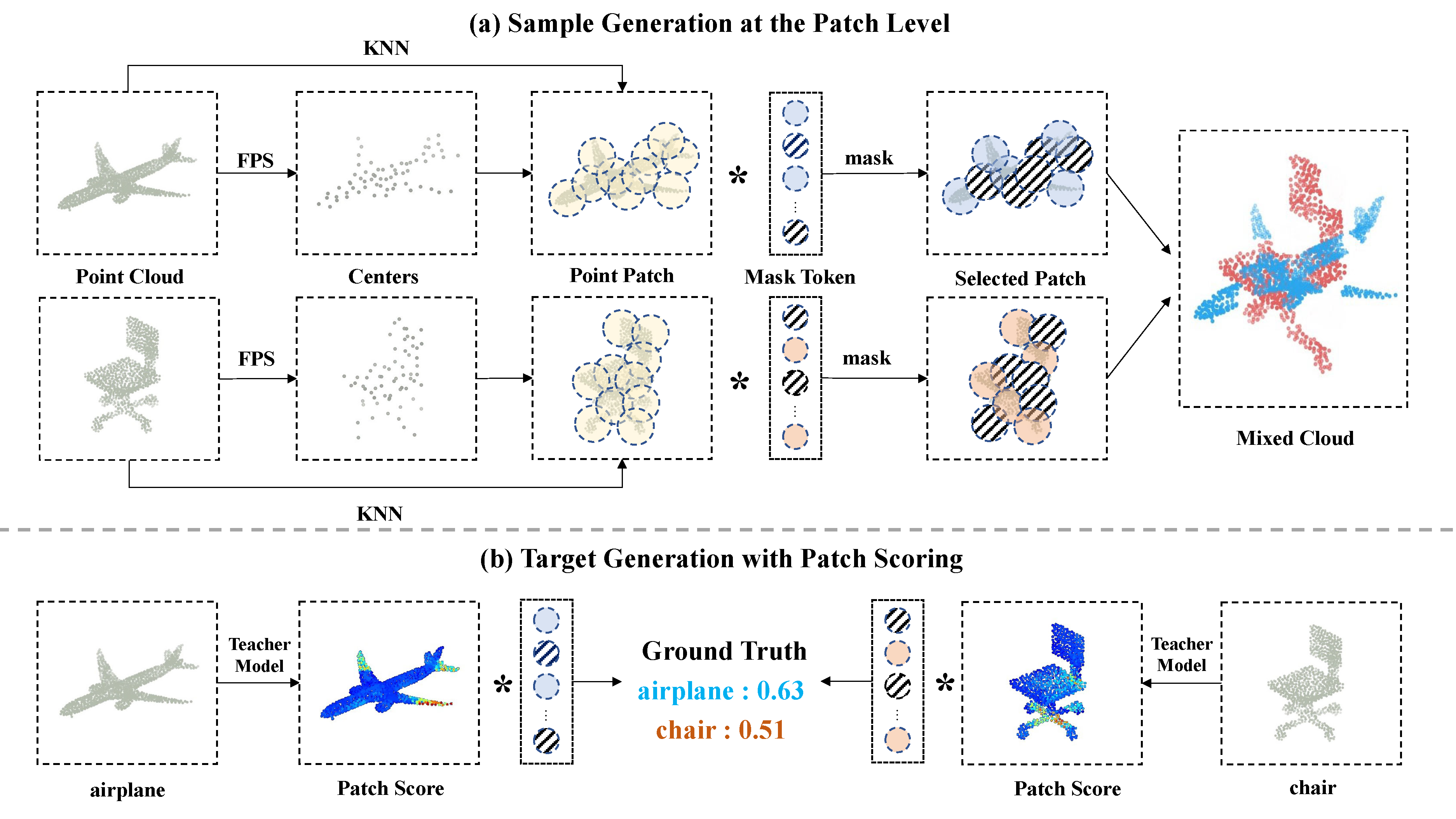}
    \caption{\textbf{The overall scheme of the PointPatchMix.} (a) The original point clouds are divided into multiple patches, which are processed by mask token and then mixed. (b) The pre-trained teacher model assigns each patch with a content-based significance score, and the ground truth of the mixed point cloud is obtained by summing the scores of selected patches.}
    \label{fig:main}
    \vspace{-8pt}
\end{figure*}

\noindent\textbf{Data augmentation on point cloud.} 
Many of the previous works~\cite{phan2018dgcnn,qi2017pointnet,qi2017pointnet++,hu2020rscnn} have mostly applied simple data augmentation at the point level, such as random jitter, rotation, and scaling. 
Li et al.~\cite{li2020pointaugment} proposed an automatic augmentation framework that optimizes the enhancer and classifier networks using adversarial learning, allowing the framework to generate more complex point cloud samples. 
PointMixUp, proposed by Chen et al.~\cite{chen2020pointmixup}, extends Mixup into point clouds to fuses two samples using optimal linear interpolation. However, the interpolation-generated samples are prone to deformation, especially for the loss of key local features. 
To keep the structural information of point clouds,  Lee et al.~\cite{lee2021regularization} proposed RsMix, which inherits the advantage of CutMix. It utilizes rigid transformations to mix two samples while maintaining the shape of the original data. 
Another related work is the PointWOLF~\cite{kim2021point}, which compensates for the scarcity of data sets by performing multiple weighted local transformations, as proposed by Kim et al. 
Meanwhile, with the application of Transformer in point clouds (such as Point-MAE~\cite{pang2022masked}), patch-level local point clouds are more effective in improving model performance than traditional points~\cite{sheshappanavar2021patchaugment}. 
In response to this requirement, we propose a new data augmentation called \emph{PointPatchMix}, which mixes point clouds at the patch level and gives a more realistic ground truth.

%% file: method.tex
\section{Methods}
In this section, we first provide a brief review of the problem setting and general process of PointCutMix~\cite{zhang2022pointcutmix}. We then introduce a novel Point Patch Scoring module, which leverages a pre-trained teacher model to generate a content-based target score. Additionally, we present an optimal assignment algorithm that assigns point clouds at the patch level. Finally, we propose PointPatchMix, a point cloud augmentation technique that operates at the token level and generates more accuracy targets.

\begin{figure*}[t]
    \centering
        \includegraphics[width=\linewidth]{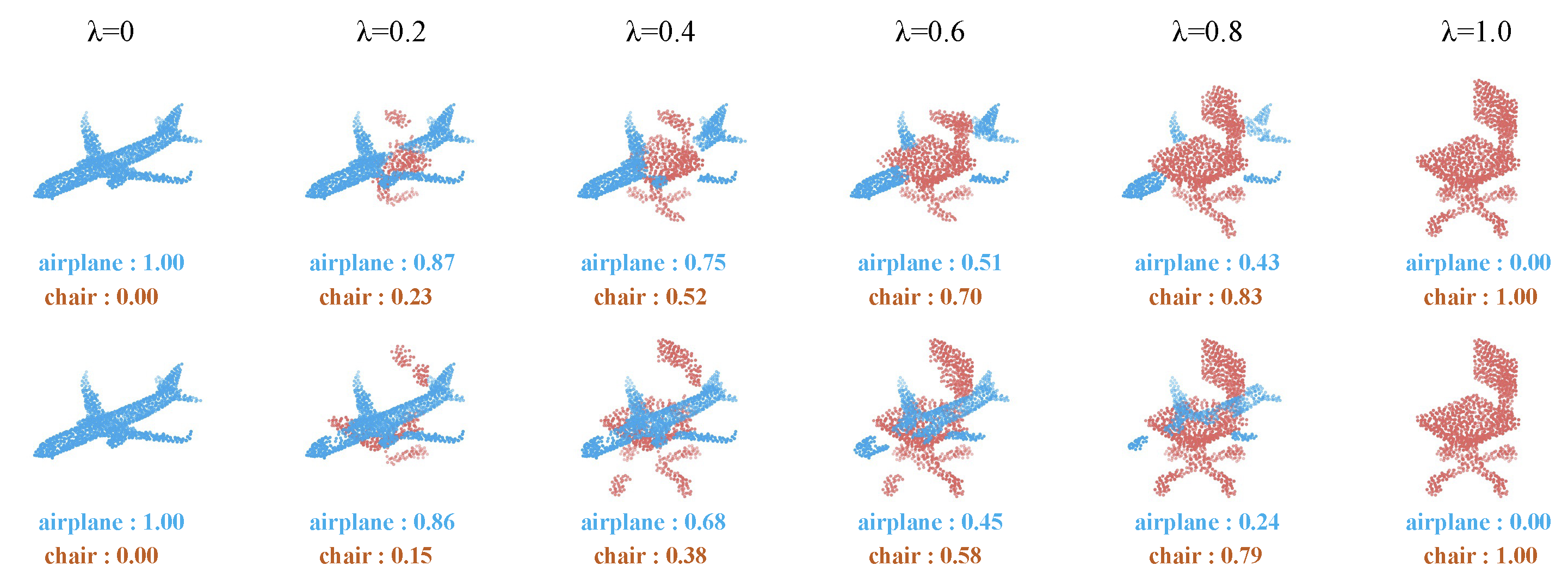}
    \caption{\textbf{The visualization of the mixed samples between a plane and a chair under different replacement ratios $\lambda$.} Top: block-level mixing. Bottom: patch-level mixing.}
    \label{fig:ratio}
    \vspace{-10pt}
\end{figure*}
\subsection{Preliminary}
\noindent\textbf{Problem Setting.}
In a standard point cloud classification task, the target is to train a function $f: x\rightarrow[0,1]^C$ to map a point cloud $x$ containing $N$ points to a one-hot class label, where $C$ is the number of whole classes, $x\in R^{3\times N}$. 
We minimize the loss between the ground truth $y$ and prediction $f(x)$ to train the network parameters.

\noindent\textbf{Revisiting PointCutMix.}
PointCutMix is proposed to mix pairs of samples with a random binary mask to alleviate the insufficient scale of point cloud datasets.
PointCutMix creates new training point cloud $(\tilde{x}, \tilde{y})$  based on a pair of point clouds $(x_{1}, y_{1})$ and $(x_{2}, y_{2})$.
The combining operation of PointCutMix is as follows:
\begin{align}
\tilde{x} & = M\odot x_{1}+(1-M)\odot x_{2} \\
\tilde{y} & = \lambda  y_{1}+(1-\lambda )y_{2} \\
\lambda & = \frac{\sum M}{N}  
\end{align}
where, $M\in\{0,1\}^{N}$ denotes which sample the point belongs to, $\odot$ denotes element-wise multiplication, and $\lambda$ is sampled from a beta distribution $Beta(\beta,\beta)$, which means $\lfloor\lambda N\rfloor$ points are selected from $x_{1}$ and $N-\lfloor\lambda N\rfloor$  points are selected from $x_{2}$. 
PointCutMix applies a similar approach to CutMix and PointMixup by creating a mixed target for the generated point clouds through a linear combination of $y_{1}$ and $y_{2}$ with $\lambda$.

We argue that block-level mixing in PointCutMix (PointCutMix-K) may not be the optimal choice for mixing-based augmentation. This is because it randomly samples one point and its nearest neighbors to mask, which restricts the diversity of the generated point clouds. On the other hand, point-level mixing (PointCutMix-R) randomly selects points and may not retain the local characteristics of the point cloud. Hence, it is necessary to balance the richness of the generated point cloud while preserving its local characteristics. Drawing inspiration from the architecture of Transformer~\cite{han2021transformer}, we propose patch-level augmentation as a viable solution for mixing-based augmentation.
In addition, the label of the mixed point cloud in PointCutMix is generated through a linear combination of the labels of the original pairs, with the mixing ratio $\lambda$ estimated solely based on the mask size. However, this simplistic approach may not be suitable for many scenarios, as the importance of each patch in the point cloud may vary significantly. Therefore, it is crucial to develop an augmentation method at the patch level that generates content-based targets.

\subsection{Point Patch Scoring}

To address the limitations of previous methods, we propose a point patch scoring module that generates content-based targets. Our intuition is that not all patches in point clouds contribute equally to classification accuracy, and we aim to balance the richness of the generated point clouds with their local characteristics by focusing on patch-level mixing. Consequently, a crucial aspect of our approach is determining each patch's significance score. A natural approach is to utilize a pre-trained teacher network based on the Transformer architecture~\cite{pang2022masked} to generate the patch significance score.

In a typical self-attention layer of a Transformer~\cite{vaswani2017attention}, the input tokens $I \in R^{N \times d}$ are used to compute the queries $Q \in R^{N \times d}$, keys $K \in R^{N \times d}$, and values $V \in R^{N \times d}$.
The attention matrix $A$ can be obtained by the dot product of queries and keys, which is then scaled by $\sqrt{d}$:
\begin{align}
A = Softmax(QK^{T}/\sqrt{d})
\end{align}

Each row of the attention matrix $A$ sums up to 1 due to the Softmax function. Then the output tokens $O$ are computed by the combination of values weighted by the attention weights:
\begin{align}
O = AV
\end{align}

\noindent Each row of matrix $A$ corresponds to the attention weights associated with a particular output token, indicating the relative contributions of all input tokens to that specific output token. Specifically, the attention weights in row $A_{1,:}$ represent the classification token, whereby $A_{1,j}$ signifies the relevance of input token $j$ to the output classification token. To facilitate pruning of the attention matrix $A$, we utilize the weights $A_{1,2},...,A_{1,N+1}$ as significance scores while disregarding $A_{1,1}$ since we retain the classification token. As the output tokens $O$ are dependent on both attention matrix $A$ and values $V$, we incorporate the norm of $V_{j}$ when determining the significance score of token $j$. This is motivated by the notion that values with a norm near zero have a negligible impact and thus, their corresponding tokens are deemed less significant. As such, the significance score of patch token $j$ is calculated as follows:
\begin{align}
S_{j}=\frac{A_{1,j}\times \left \| V_{j} \right \| }{ {\textstyle \sum_{i=2}A_{1,i}\times\left \| V_{i} \right \|  } }
\end{align}
\noindent where $i,j \in {2...N}$. For a multi-head attention layer, we compute the significance scores for each head separately and subsequently aggregate them by taking the sum over all heads.

\subsection{Point Patch Mixing.}
\noindent\textbf{Optimal patch assignment.}
Mixing patches at the patch level requires establishing a one-to-one relationship between the patches of two input point clouds. In the image domain, pixels are arranged in a grid formation, and two images can be naturally matched according to their coordinates by resizing or cropping them to the same size. However, point clouds lack a predetermined order and are permutation-invariant, making it essential to define the correspondence between patches based on rules other than their position.

Following PointMixup~\cite{chen2020pointmixup} and PointCutMix~\cite{zhang2022pointcutmix}, we adopt the Earth Mover's Distance (EMD) function to establish an optimal assignment between two point clouds, due to its efficacy in accounting for the geometric relationship, local details, and density distributions of two point clouds. The point clouds are denoted as $x_{1}$ and $x_{2}$. The EMD calculates the minimum total displacement required for matching each point in $x_{1}$ to the corresponding point in $x_{2}$. The assignment function $\theta^{*}$ in the EMD can be calculated as:
\begin{align}
\label{equ:1}
\theta^{*}= argmin_{\theta \in \Theta} \sum_{i} \left \| x_{1,i}-x_{2,\theta(i)} \right \| _{2}  
\end{align}

\noindent where $\Theta$ gives a one-to-one correspondences between two point clouds at the point level.
And getting the assignment function at the point level. We can easily extend it into the patch level. 
\begin{align}
\label{equ:2} 
\theta^{*}_P= argmin_{\theta \in \Phi } \sum_{p}  \sum_{i \in p}(\left \|x_{1,i}-x_{2,\theta(i)}\right \| _{2} )  
\end{align}

\noindent where $\Phi$ give a one-to-one correspondeces between two point clouds at the patch level, $p$ is the number of patches.
We can use the total points to calculate the patch correspondences with the lowest EMD at the patch level by Equation \ref{equ:2}. What's more, we find that directly using the center point position of each patch to represent the whole patch and Equation \ref{equ:1} can obviously save the preprocessing time and get a similar performance. Thus we use the center points of each patch to calculate the optimal patch assignment.

\noindent\textbf{Mixing algorithm.} 
Figure \ref{fig:main} presents an overview of our PointPatchMix approach. Firstly, we divide the input point cloud $x$ into patches $x_{p}$. Next, we generate a patch-level random mask $M^t$ based on the mask-out ratio $\lambda$. With the significance score calculated by the pre-trained teacher model. The mixed new training sample $f$ is then constructed as follows:
\begin{align}
\tilde{x}^{p} & = M^t\odot x^{p}_{1}+(1-M^t)\odot x^{p}_{2}\\
\tilde{y} & = \textstyle \sum_{j=2} M^t_{j}\odot S_{1j}+\sum_{j=2}(1-M^t_{j})\odot S_{2j}
\end{align}

\noindent where $j \in {2...N}$, $\odot$ denotes element-wise multiplication, $M^t_{j}$ denotes the $j$-th patch token of the mask $M^t$, $S_{1j}$ and $S_{2j}$ are the $j$-th patch token of the significance score of $x_1$ and $x_2$ respectively. The visualization of mixed samples under different replacement ratios are shown in Figure \ref{fig:ratio}.

\begin{table}[]
\caption{\textbf{Comparision with state-of-the-art methods on ScanObjectNN.} We report the classification accuracy (\%) on three splits of ScanObjectNN.}
\label{table:scan}
\resizebox{\linewidth}{!}{%
\begin{tabular}{cccc}
\hline
Methods & OBJ-BG & OBJ-ONLY & PB-T50-RS \\ \hline
PointNet~\cite{qi2017pointnet} & 73.3 & 79.2 & 68 \\
SpiderCNN~\cite{xu2018spidercnn} & 77.1 & 79.5 & 73.7 \\
PointNet++~\cite{qi2017pointnet++} & 82.3 & 84.3 & 77.9 \\
DGCNN~\cite{phan2018dgcnn} & 82.8 & 86.2 & 78.1 \\
PointCNN~\cite{li2018pointcnn} & 86.1 & 85.5 & 78.5 \\
BGA-PN++~\cite{uy2019revisiting} & - & - & 80.2 \\
GBNet~\cite{qiu2021geometric} & - & - & 80.5 \\
PRANet~\cite{cheng2021net} & - & - & 81 \\
Transformer~\cite{yu2022point} & 79.8 & 80.5 & 88.2 \\
Transformer-OcCo~\cite{yu2022point} & 84.8 & 85.5 & 78.8 \\
Point-BERT~\cite{yu2022point} & 87.4 & 88.1 & 83 \\ \hline
Point-MAE~\cite{pang2022masked} & 90.0 & 88.2 & 85.1 \\
+PointPatchMix & \textbf{90.9} & \textbf{91.0} & \textbf{86.3} \\ \hline
\end{tabular}
}
\vspace{-8pt}
\end{table}

\begin{table}[]
\centering
\caption{\textbf{Comparision with state-of-the-art methods on ModelNet40.} We compare our approach with various self-supervised and supervised methods. JRST represents jitter, random rotation, random scaling, and translation respectively.}
\label{table:modelnet}
\vspace{-8pt}
\begin{minipage}[c]{\linewidth}
\centering
\subcaption{Supervised methods}
\begin{tabular}{ccc}
\hline
Methods & Augmentation & Accuracy \\ \hline
PointNet~\cite{qi2017pointnet} & JRST & 89.2 \\
PointNet++~\cite{qi2017pointnet++} & JRST & 90.7 \\
PointCNN~\cite{li2018pointcnn} & T & 92.5 \\
KPConv~\cite{thomas2019kpconv} &  color drop & 92.9 \\
DGCNN~\cite{phan2018dgcnn} & ST & 92.9 \\
RS-CNN~\cite{hu2020rscnn} & ST & 92.9 \\
PCT~\cite{guo2021pct} & point drop, T & 93.2 \\
PVT~\cite{zhang2108pvt} & JRST & 93.6 \\
PointTransformer~\cite{zhao2021point} & color auto-contrast & 93.7 \\ \hline
PointMLP~\cite{ma2022rethinking} & T & 94.5 \\
+PointPatchMix & PointPatchMix & \textbf{94.7} \\ \hline
\end{tabular}
\end{minipage}
\begin{minipage}[c]{\linewidth}
\centering
\subcaption{Self-supervised methods}
\begin{tabular}{ccc}
\hline
Methods  & Augmentation & Accuracy \\ \hline
OcCo~\cite{wang2021unsupervised}  & ST  & 93.0 \\
STRL~\cite{huang2021spatio}  & T & 93.1 \\
IAE~\cite{yan2022implicit}  & ST & 93.7 \\
Transformer-OcCo~\cite{yu2022point}  & ST & 92.1 \\
Point-BERT~\cite{yu2022point}  & resampling  & 93.2 \\ \hline
Point-MAE~\cite{pang2022masked}  & ST & 93.8 \\
+PointPatchMix  & PointPatchMix & \textbf{94.1} \\ \hline
\end{tabular}
\end{minipage}
\vspace{-8pt}
\end{table}


%% file: experiments.tex
\section{Experiments}

\begin{table*}[ht]
\centering
\caption{\textbf{Few-shot classification on ModelNet40.} We report the accuracy with standard of 10 independent experiments.}
\label{table:few}
\begin{tabular}{ccccc}
\hline
Methods & 5-way,10-shot & 5-way,20-shot & 10-way,10-shot & 10-way,20-shot \\ \hline
PointNet~\cite{qi2017pointnet} & 52.0 $\pm$ 3.8 & 57.8 $\pm$ 4.9 & 46.6 $\pm$ 4.3 & 35.2 $\pm$ 4.8 \\
PointNet + OcCo~\cite{wang2021unsupervised} & 89.7 $\pm$ 1.9 & 92.4 $\pm$ 1.6 & 93.9 $\pm$ 1.8 & 89.7 $\pm$ 1.5 \\
PointNet + CrossPoint~\cite{afham2022crosspoint} & 90.9 $\pm$ 4.8 & 93.5 $\pm$ 4.4 & 83.6 $\pm$ 4.7 & 90.2 $\pm$ 2.2 \\
DGCNN~\cite{phan2018dgcnn} & 31.6 $\pm$ 2.8 & 40.8 $\pm$ 4.6 & 19.9 $\pm$ 2.1 & 16.9 $\pm$ 1.5 \\
DGCNN + OcCo~\cite{wang2021unsupervised}& 90.6 $\pm$ 2.8 & 92.5 $\pm$ 1.9 & 82.9 $\pm$ 1.3 & 86.5 $\pm$ 2.2 \\
DGCNN + CrossPoint~\cite{afham2022crosspoint} & 92.5 $\pm$ 3.0 & 94.9 $\pm$ 2.1 & 83.6 $\pm$ 5.3 & 87.9 $\pm$ 4.2 \\
Transformer-rand~\cite{yu2022point} & 87.8 $\pm$ 5.2 & 93.3 $\pm$ 4.3 & 84.6 $\pm$ 5.5 & 89.4 $\pm$ 6.3 \\
Transformer-OcCo~\cite{yu2022point} & 94.0 $\pm$ 3.6 & 95.9 $\pm$ 2.3 & 89.4 $\pm$ 5.1 & 92.4 $\pm$ 4.6 \\
Point-BERT~\cite{yu2022point} & 94.6 $\pm$ 3.1 & 96.3 $\pm$ 2.7 & 91.0 $\pm$ 5.4 & 92.7 $\pm$ 5.1 \\ \hline
Point-MAE~\cite{pang2022masked} & 96.3 $\pm$ 2.5 & 97.8 $\pm$ 1.8 & 92.6 $\pm$ 4.1 & 95.0 $\pm$ 3.0 \\
\textbf{+PointPatchMix} & \textbf{96.6 $\pm$ 2.2} & \textbf{97.8 $\pm$ 1.7} & \textbf{92.7 $\pm$ 3.3} & \textbf{95.2 $\pm$ 2.7} \\ \hline
\end{tabular}
\label{table}
\vspace{-8pt}
\end{table*}

\subsection{Datasets}
We conduct extensive experiments on both synthetic and real-world datasets in point cloud shape classification to evaluate the effectiveness of PointPatchMix, i.e., ModelNet40~\cite{wu20153d} and ScanObjetNN~\cite{uy2019revisiting}.

\noindent\textbf{ModelNet40.}   
It is a widely-used clean point cloud object dataset for point cloud classification tasks, comprising 12,311 samples spanning 40 object categories.
We follow the standard pattern, using 9843 samples for training and 2468 samples for testing.
During the training process, we exclusively utilize the proposed hybrid method for data augmentation. 
For a fair comparison, we adopt the standard voting methods~\cite{liu2019relation} used in prior work during the testing process.

\noindent\textbf{ScanObjectNN.} 
It is a point-cloud object dataset derived from the real world, which contains about 15,000 samples of 15 object categories. 
These samples are obtained from real scenes, often with occlusion and noise, and therefore extremely challenging. No voting method is used during testing on ScanObjectNN.

\subsection{Experimental Details}
To ensure that PointPatchMix is a general data augmentation method, we select multiple popular architectures in the current point cloud field to evaluate its availability and effectiveness in various network architectures. 
In the specific experimental configuration, all networks are uniformly provided 1024 points for training learning with 300 epochs and a batch size of 32. 
Meanwhile, we maintain the same configuration as possible with the original published paper to facilitate a fair comparison with the baseline. 
For PointNet, PointNet++, we use the Adam optimizer with an initial learning rate of 0.001 and a decay rate of 0.5 per 20 cycles. 
For Point-MAE, we use the AdamW optimizer with an initial learning rate of 0.001 and a weight decay of 0.05.A cosine annealing strategy is used to attenuate the learning rate.
We adopt Point-MAE as our teacher model to generate the patch token scores and save the scores offline to improve training efficiency.

\subsection{Experimental Results}
In this subsection, we utilize Point-MAE, a robust classification model, as our baseline. PointPatchMix achieves impressive improvements in various scenarios, including real-world data, synthetic data, and few-shot learning settings.

\noindent\textbf{Comparision with state-of-the-art methods on ScanObjectNN.}
Table \ref{table:scan} demonstrates the classification results on ScanObjectNN. We use the Point-MAE as our baseline model and set a new state-of-the-art on all the settings. Our PointPatchMix exceeds Point-MAE by an improvement of 0.9\% on OBJ-BG, 2.8\% on OBJ-ONLY, and 1.2\% on PB-T50-RS, which is a very significant improvement on the challenging real-world dataset.

\noindent\textbf{Comparision with state-of-the-art methods on ModelNet40.}
Table \ref{table:modelnet} presents the classification results on ModelNet40.
All the reported methods are given 1024 points that only contain coordinate information without any normal information.
Our PointPatchMix achieves 94.1\% accuracy, and even outperforms Point-MAE with 8192 points as input.
We also adopt PointMLP as our baseline and PointPatchMix further improve the accuracy to 94.7\%, thus the effectiveness of our PointPatchMix for point-based architecture can be proved.

\noindent\textbf{Few-shot Learning.}
In order to assess the generalization ability of PointPatchMix with limited training data, we conduct the few-shot classification task on ModelNet40. Following standard practice, we performed $N$-way $K$-shot experiments on PointPatchMix 10 times, randomly selecting $N$ classes from ModelNet40 and sampling $K$ objects from each class. The results are presented in Table \ref{table:few}, and demonstrate that PointPatchMix outperforms previous methods in all few-shot settings, indicating its strong capacity with limited training data.
\begin{table}[]
\centering
\caption{\textbf{Comparision with other mixing-based methods.} We report the classification accuracy(\%).}
\label{table:mixing}
\resizebox{\linewidth}{!}{%
\begin{tabular}{cccc}
\hline
Method & PointNet & PointNet++ & Transformer  \\ \hline
baseline & 89.2 & 90.7 & 91.4 \\
PointMixup~\cite{chen2020pointmixup} & 89.9 & 92.7 & 92.3  \\
RSMix~\cite{lee2021regularization} & 89.7 & 92.1 & 92.5  \\
PointCutMix~\cite{zhang2022pointcutmix} & 89.6 & \textbf{93.4} & 92.5  \\
\textbf{PointPatchMix} & \textbf{90.1} & 92.9 & \textbf{93.3}  \\ \hline
\end{tabular}
}
\vspace{-8pt}
\end{table}

\begin{table}[]
\centering
\caption{\textbf{Ablation Studies.} We conducted experiments in a total of four perspectives: mixing level, target generation, patch assignment, and the influence of $\beta$, and report the classification accuracy (\%) of Transformer on ModelNet40.}
\label{table:ablation}
\resizebox{\linewidth}{!}{%
\begin{tabular}{ccccc}
\hline
\begin{tabular}[c]{@{}c@{}}Mixing\\ Level\end{tabular} & \begin{tabular}[c]{@{}c@{}}Target\\ Generation\end{tabular} & \begin{tabular}[c]{@{}c@{}}Patch \\ Assignment\end{tabular} & \begin{tabular}[c]{@{}c@{}}Beta\\ Value\end{tabular} & Accuracy \\ \hline
- & - & - & - & 91.4 \\
Patch & Linear & All & 1 & 92.4 \\
Patch & Score & All & 1 & 92.9 \\
Block & Score & All & 1 & 91.9 \\
Patch & Score & Center & 1 & 92.9 \\
Patch & Score & Center & 0.5 & 92.8 \\
Patch & Score & Center & 1.5 & \textbf{93.3} \\
Patch & Score & Center & 2 & 93.1 \\ \hline
\end{tabular}
}
\vspace{-8pt}
\end{table}

\subsection{Ablation Studies}
To evaluate the contribution of each component in PointPatchMix, we conduct ablation studies on ModelNet40.
We mainly use the Transformer proposed in Point-MAE\cite{pang2022masked} without pre-training as our baseline model for evaluation, which archives 91.4\% on ModelNet40.

\noindent\textbf{Comparision with other mixing-based methods on ModelNet40.}
We compare PointPatchMix with other mixing-based methods in Table \ref{table:mixing}. As shown in the table, PointPatchMix achieves better performance with both PointNet and Transformer.
We can observe that other mixing-based methods improve limited with Transformer compared to PointNet and PointNet++. On the contrary, our proposed PointPatchMix improves our baseline by 1.9\% and improves PointCutMix by 0.8\% with Transformer, which validates that PointPatchMix fits well for Transformer-based architectures. 

\noindent\textbf{Mixing Level.} In our ablation study, we compared patch-level mixing with block-level mixing and observed that patch-level mixing outperforms block-level mixing by 1.0\% on ModelNet40, as shown in Table \ref{table:ablation}. This indicates that selecting points at the patch level generates more diverse training data while preserving the local features of the point clouds. Hence, we chose patch-level mixing as our main approach for PointPatchMix.

\noindent\textbf{Target Generation.}
To validate the effectiveness of our proposed point patch scoring module, we compared it with the approach of using a linear combination to generate the targets based on the number of selected patches for each point cloud. Our experimental results clearly demonstrate that our content-based point patch scoring method outperforms the linear combination method, indicating that our approach can better capture the content information of each object and generate more accurate targets.

\noindent\textbf{Optimal patch assignment points.}
We also explore two different ways of patch assignment, and we find that there are no obvious differences between using all points and using only center points for optimal patch assignment, while center points based patch assignment requires less preprocessing time. Thus using center points for optimal patch assignment is a more efficient and practical option.

\noindent\textbf{Influence of $\beta$.}
We investigate the influence of $\beta$ which is used to sample the data from a beta distribution. Our model achieves best when $\beta=1.5$. Thus we set $\beta=1.5$ as default for all experiments.

\begin{figure*}[t]
    \centering
        \includegraphics[width=\linewidth]{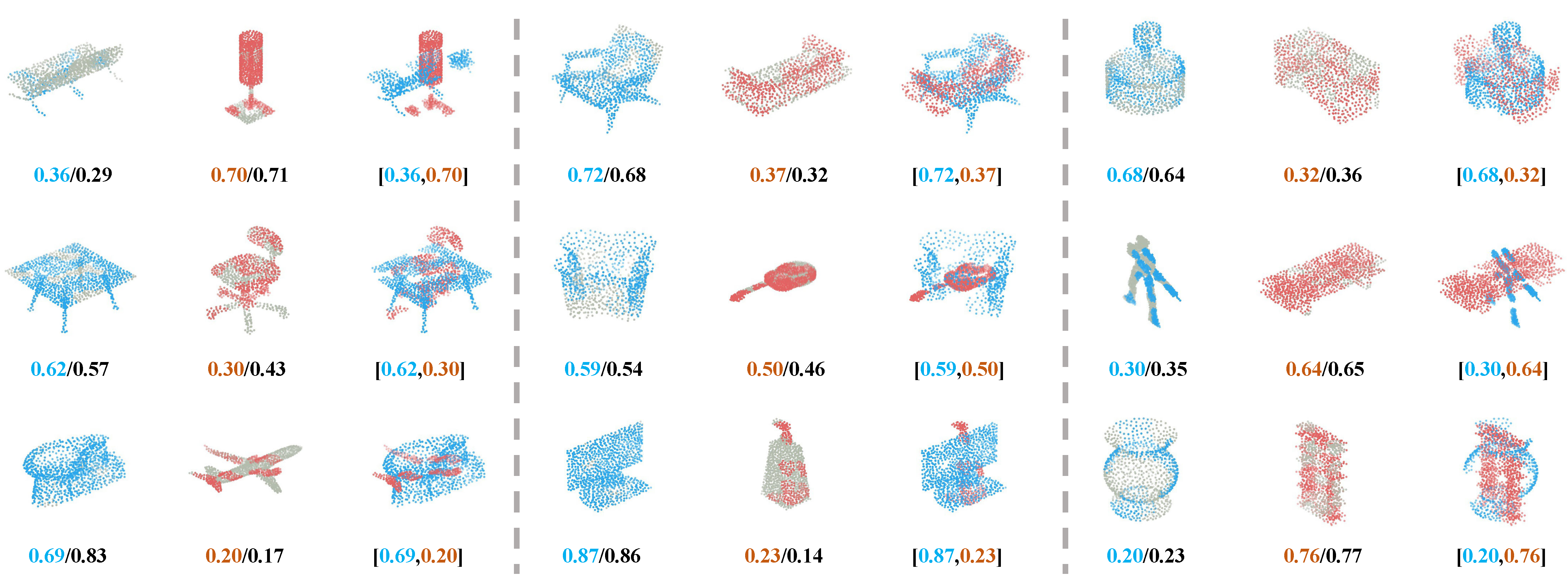}
    \caption{\textbf{Qualitative examples of PointPatchMix.} For each set of samples, the mixed point cloud (right) consists of blue and red patches, where the blue patches are randomly selected from the left object and the red patches are the corresponding complementary part from the middle object. At the bottom of the point clouds, the blue and red numbers represent the scores of these patches, while the black numbers are the percentage of the number of points.}
    \label{fig:mix}
    \vspace{-5pt}
\end{figure*}

\begin{figure*}[t]
    \centering
        \includegraphics[width=\linewidth]{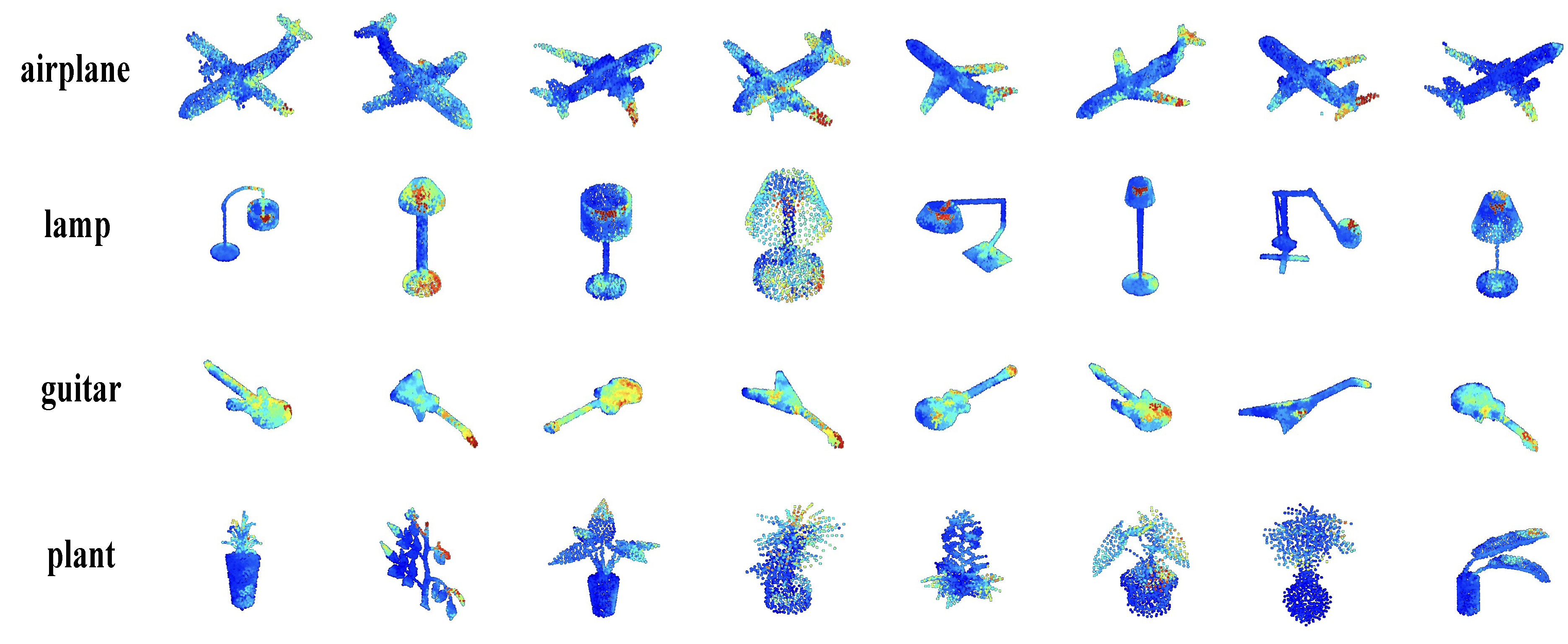}
    \caption{\textbf{Patch scores for different types of point clouds.} Content-rich patches are closer to red, while unimportant patches are closer to blue. As we can directly observe, edge patches (e.g., the wing of airplane, the head of guitar) and some key local patches (e.g., the wick of lamp, the leaf of plant) often contain rich semantics.}
    \label{fig:attention}
    \vspace{-5pt}
\end{figure*}

\noindent\textbf{Robustness.}
We tested the robustness~\cite{sun2022benchmarking} of PointPatchMix with Transformer to four noisy environments: jitter, rotation, scaling, and DropPoint, in order to verify that our method makes the model robust to noise.
As we can see from Table \ref{table:robustness}, PointPatchMix improves the baseline model notably on all the test settings, which verified that PointPatchMix can successfully improve the robustness ability of the baseline model.

\begin{table}[]
\caption{\textbf{Robustness analysis.} We report the classification accuracy (\%) with Transformer to four noisy environments: jitter, rotation, scaling, and DropPoint.}
\label{table:robustness}
\centering
\begin{tabular}{ccccc}
\hline
Transforms & Transformer  & +PointPatchMix \\ \hline
- & 91.4 & 93.3 (1.9$\uparrow$) \\
Noise $\sigma$=0.01 & 91.0 & 92.3 (1.3$\uparrow$)  \\
Noise $\sigma$=0.03 &  75.8  & 79.7 (3.9$\uparrow$)  \\
Z-rotation{[}-30 30{]} & 80.2 & 85.2 (5.0$\uparrow$)  \\
X-rotation{[}-30 30{]} & 81.7 & 87.6 (5.9$\uparrow$)  \\
Y-rotation{[}-30 30{]} & 87.5 & 90.5 (3.0$\uparrow$)  \\
Scale (0.6) & 68.3  & 81.7 (14.4$\uparrow$)  \\
Scale (2.0) &  42.3& 47.4 (5.1$\uparrow$)  \\
DropPoint(0.2) & 91.1  & 92.5 (1.4$\uparrow$)  \\ \hline
\end{tabular}
\vspace{-8pt}
\end{table}

\subsection{Qualitative Analysis}

\noindent\textbf{Qualitative examples of PointPatchMix.} We present additional examples of PointPatchMix for qualitative analysis in Figure \ref{fig:mix}. The diversity of target values and patch ratios in the generated point clouds highlights the effectiveness of our proposed patch scoring module. By generating targets based on patch contents and emphasizing important regions during training, PointPatchMix enables the network to focus on informative areas, resulting in improved point cloud classification performance. 

\noindent\textbf{Visualization of Patch Scores.} As previously discussed, traditional methods for determining the targets of mixed point clouds rely on linear combinations of labels from the paired mixing point clouds, which can be imprecise as the significance of each patch may vary. To mitigate this issue, we propose a novel point patch scoring module that allows the baseline model to concentrate on informative areas, resulting in improved performance. We present a visualization of the patch scores in Figure \ref{fig:attention}, which illustrates that informative areas such as the edges have higher scores and can lead to consistent performance gain by directing the baseline's attention to these areas.

%% file: conclusion.tex
\section{Conclusion}
Data augmentation is a critical and challenging task in point cloud processing. To address this issue, we propose a novel mixing-based data augmentation method, PointPatchMix. This approach generates diverse training samples while preserving the local characteristics of point clouds. Our patch-level mixing strikes a balance between block-based and point-level mixing methods and outperforms their performance. Furthermore, our patch scoring module assigns more realistic content-based targets to the mixed point clouds. Experimental results demonstrate that PointPatchMix performs competitively in various settings, including real-world point cloud classification, synthetic point cloud classification, and few-shot learning settings. It also demonstrates strong generalization across multiple architectures and enhances the baseline model's robustness.

Our study focuses primarily on point cloud classification, which is crucial in 3D scene understanding. However, point cloud data has various applications in other domains, such as scene and part segmentation. Thus, a promising future direction would be to broaden the scope of our investigation to encompass these areas. This effort could contribute to the development of more comprehensive and sophisticated point augmentation methods for point cloud analysis.